\documentclass[conference]{IEEEtran}
\IEEEoverridecommandlockouts
\usepackage[utf8]{inputenc}
\usepackage{cite}

\usepackage{comment}
\usepackage{placeins}
\usepackage{amsmath,amssymb,amsfonts}
\usepackage{algorithmic}
\usepackage{tabularx,tabulary}
\usepackage{graphicx}
\usepackage{textcomp}
\usepackage{xcolor}
\usepackage[flushleft]{threeparttable}
\usepackage[inline]{enumitem}
\def\BibTeX{{\rm B\kern-.05em{\sc i\kern-.025em b}\kern-.08em
    T\kern-.1667em\lower.7ex\hbox{E}\kern-.125emX}}
\newlength\mylength
\begin{document}


\title{Drones on the Rise: Exploring the Current and Future Potential of UAVs\\}

\author{S. M.~Riazul~Islam
\thanks{S. M. Riazul Islam (s.mr.islam@hud.ac.uk) is with the Department of Computer Science, University of Huddersfield, Huddersfield, HD1 3DH, UK.}}

\maketitle

\begin{abstract}
Unmanned Aerial Vehicles (UAVs) have become increasingly popular in recent years due to their versatility and affordability. This article provides an overview of the history and development of UAVs, as well as their current and potential applications in various fields. In particular, the article highlights the use of UAVs in aerial photography and videography, surveying and mapping, agriculture and forestry, infrastructure inspection and maintenance, search and rescue operations, disaster management and humanitarian aid, and military applications such as reconnaissance, surveillance, and combat. The article also explores potential advancements in UAV technology and new applications that could emerge in the future, as well as concerns about the impact of UAVs on society, such as privacy, safety, security, job displacement, and environmental impact. Overall, the article aims to provide a comprehensive overview of the current state and future potential of UAV technology, and the benefits and challenges associated with its use in various industries and fields.
\end{abstract}


\section{Introduction}
UAV stands for Unmanned Aerial Vehicle, which is a type of aircraft that operates without a human pilot on board. UAVs can be controlled remotely by a human operator, or can be programmed to fly autonomously using pre-defined routes and waypoints. UAVs come in a variety of sizes, ranging from small hand-held models to large military drones used for surveillance and combat missions \cite{vinogradov2019tutorial}. They are used for a variety of purposes, including aerial photography, surveying and mapping, agricultural monitoring, search and rescue operations, and military reconnaissance. A brief history of UAV development:

1917: During World War I, the first UAVs were developed by the US and British militaries for use as aerial targets.

1930s: The British Royal Air Force developed the first true UAV, called the Queen Bee, which was a remote-controlled aircraft used for training purposes.

1950s: The US military began developing UAVs for reconnaissance and surveillance purposes, including the Ryan Model 147, which was used during the Vietnam War.

1970s: Advances in microelectronics and miniaturization led to the development of smaller, more sophisticated UAVs, such as the Israeli Scout and the US Pioneer.

1990s: UAV technology continued to improve, and the US military began using UAVs in combat situations, including the Persian Gulf War and the Balkan conflicts.

2000s: The use of UAVs expanded beyond military applications, with civilian agencies and industries beginning to use UAVs for tasks such as aerial photography, surveying and mapping, and environmental monitoring.

Present day: UAV technology continues to advance rapidly, with improvements in areas such as battery life, range, and autonomous capabilities. UAVs are being used in an ever-widening range of applications, from package delivery to wildlife conservation to disaster response.

There are many reasons for the growing importance of UAVs \cite{gupta2015survey, rovira2022review}. Some of the key reasons are outlined as follows:

\textbf{Enhanced Efficiency:} UAVs can perform many tasks faster and more efficiently than traditional methods, such as aerial photography, mapping, and surveying. UAVs can also be used in hazardous or hard-to-reach areas, reducing risk to human personnel.

\textbf{Cost-effectiveness:} UAVs can be less expensive than traditional aircraft, and can perform tasks at a fraction of the cost of manned flights. This makes them particularly useful for smaller companies or organizations that cannot afford the cost of manned aircraft.

\textbf{Versatility:} UAVs can be used in a wide range of applications, from military and surveillance operations to agriculture and wildlife conservation. They can also be equipped with a variety of sensors and cameras, making them adaptable to many different types of tasks.

\textbf{Increased Safety:} UAVs can be used in situations where human safety is a concern, such as in search and rescue operations, disaster response, and infrastructure inspection. They can also be used to monitor dangerous or hazardous environments, such as oil rigs, without putting human personnel at risk.

\textbf{Environmental Benefits:} UAVs can be used to monitor and study the environment, including tracking wildlife populations, monitoring deforestation and other environmental changes, and measuring air and water quality.

Followings are some notable UAV projects underway worldwide that demonstrate the diversity of UAV applications and the potential for UAV technology to have a positive impact on various industries and sectors.
\begin{itemize}
    
    \item
    Project Wing by Google: A drone delivery project that aims to deliver goods to customers via UAVs.

    \item
Zipline: A medical supply delivery service using UAVs to reach remote areas in Africa and other parts of the world.

\item
DJI Phantom 4 RTK: A high-precision mapping drone that can be used for surveying and construction applications.

\item
Skysense: A drone charging station system that enables UAVs to recharge autonomously.

\item
MQ-9 Reaper: A military UAV used for reconnaissance, surveillance, and combat operations.
\end{itemize}

In this article, we will provide an overview of the applications of UAVs in various industries and fields. We will also explore potential advancements in UAV technology and new applications that could emerge in the future, as well as concerns about the impact of UAVs on society, such as privacy, safety, security, job displacement, and environmental impact. 

\section{Civilian Applications of UAVs}
Civilian applications of UAVs include aerial photography and videography, surveying and mapping, agriculture and forestry, infrastructure inspection and maintenance, search and rescue operations, and various other tasks where UAVs can provide safer, more cost-effective, and efficient solutions compared to traditional methods \cite{ghamari2022unmanned}.

\subsection{Aerial Photography and Videography}
Here are some of the key applications of UAVs in aerial photography and videography \cite{li2012design}:

\textbf{Film and TV Production:} UAVs are commonly used in the film and television industry to capture aerial footage of landscapes, cityscapes, and action scenes. They can provide unique angles and perspectives that would be difficult or impossible to capture with traditional methods.

\textbf{Real Estate Marketing:} UAVs are increasingly used in the real estate industry to capture aerial footage of properties for marketing purposes. This can give potential buyers a better sense of the property and its surroundings.

\textbf{Event Coverage:} UAVs can be used to capture aerial footage of outdoor events, such as concerts, sports games, and festivals. This can provide a unique perspective on the event and create dynamic and engaging footage for promotional materials.

\textbf{Tourism and Travel:} UAVs can be used to capture aerial footage of popular tourist destinations, such as landmarks, beaches, and natural wonders. This can provide potential visitors with a better sense of the destination and attract more tourism.

\subsection{Surveying and Mapping}
Below we discuss some ways on how UAVs can be used in surveying and mapping \cite{nex2014uav, wen2021study}:

\textbf{Topographic Mapping:} UAVs can be equipped with high-resolution cameras and LiDAR (Light Detection and Ranging) sensors to capture detailed topographic maps of an area. This data can be used to create 3D models, contour maps, and elevation profiles.

\textbf{Land Surveying:} UAVs can be used to capture accurate measurements and data of a specific land area. This data can be used to create detailed land surveys for construction, engineering, and planning purposes.

\textbf{Infrastructure Inspection:} UAVs can be used to inspect critical infrastructure, such as bridges, power lines, and pipelines. This can provide detailed visual data and detect potential problems or damage.

\textbf{Agricultural Mapping:} UAVs can be used to capture aerial imagery and data of agricultural fields, such as crop health, plant density, and water management. This data can be used to optimize crop yields and reduce costs.

\textbf{Environmental Monitoring:} UAVs can be used to monitor environmental changes, such as deforestation, erosion, and natural disasters. This data can be used to create detailed environmental impact reports and monitor changes over time.

\subsection{Agriculture and Forestry}
Major applications of UAVs in agriculture and forestry are outlined below \cite{radoglou2020compilation, aslan2022comprehensive}.

\textbf{Crop Health Monitoring:} UAVs can be equipped with multispectral or hyperspectral cameras to capture images of crops in different spectral bands. This data can be used to identify crop health issues such as nutrient deficiencies, disease, and pests, which can help farmers to address these issues in a timely manner.

\textbf{Crop Mapping:} UAVs can be used to capture high-resolution images of crop fields and create detailed maps of crop yields and plant health. This data can help farmers to optimize crop management strategies, such as irrigation, fertilization, and crop rotation.

\textbf{Precision Agriculture:} UAVs can be used to provide precise data for precision agriculture techniques such as variable rate application of fertilizers and pesticides. This can reduce costs and environmental impact by only applying inputs where they are needed.

\textbf{Forest Management:} UAVs can be used to monitor forest health, map tree species and density, and detect forest fires. This data can help forest managers to optimize forest management strategies such as harvesting, replanting, and pest management.

\textbf{Wildlife Monitoring:} UAVs can be used to monitor wildlife populations and detect illegal activities such as poaching \cite{hodgson2016precision}. This can help conservationists and authorities to protect endangered species and enforce wildlife protection laws.

\subsection{Infrastructure Inspection and Maintenance}
There are several applications of UAVs in infrastructure inspection and maintenance \cite{lekidis2022electricity, 9778241}, including:

\textbf{Bridge Inspection:} UAVs can be used to inspect the structural integrity of bridges and detect any potential damage or defects. This can be done by capturing high-resolution images of the bridge from different angles, which can be analyzed by engineers.

\textbf{Pipeline Inspection:} UAVs can be used to inspect pipelines for leaks, cracks, or other damage. This can be done by capturing high-resolution images and video of the pipeline from different angles, which can be analyzed by engineers.

\textbf{Power Line Inspection:} UAVs can be used to inspect power lines for damage, corrosion, or other issues. This can be done by capturing high-resolution images and video of the power lines from different angles, which can be analyzed by engineers.

\textbf{Building Inspection:} UAVs can be used to inspect the exterior of buildings for damage, wear and tear, or other issues. This can be done by capturing high-resolution images and video of the building from different angles, which can be analyzed by engineers.

\textbf{Road and Rail Inspection:} UAVs can be used to inspect roads and rail infrastructure for damage, wear and tear, or other issues. This can be done by capturing high-resolution images and video of the infrastructure from different angles, which can be analyzed by engineers.

\subsection{Search and Rescue Operations} 
Another key area of civilian application of UAVs is found in search and rescue operations \cite{alsamhi2022uav, yanmaz2023joint} such as

\textbf{Aerial Surveillance:} UAVs can be used to quickly search large and remote areas for missing persons or other targets. They can be equipped with high-resolution cameras and thermal imaging sensors, which can provide real-time video footage to search teams on the ground.

\textbf{Communication Relay:} UAVs can be used as communication relays between rescue teams and missing persons. They can be equipped with communication equipment such as radios, which can be used to establish contact with the missing person and provide vital information to search teams.

\textbf{Delivery of Supplies:} UAVs can be used to deliver emergency supplies such as food, water, and medicine to stranded or injured individuals in remote or inaccessible areas. This can be particularly useful in areas where access by ground or air is limited.

\textbf{Hazard Assessment:} UAVs can be used to assess hazards and risks in areas that are difficult or dangerous for search teams to access. They can be equipped with sensors to detect hazardous conditions such as gas leaks, fire, or unstable terrain.

\textbf{Mapping and Documentation:} UAVs can be used to create high-resolution maps of search areas and document the search process. This can help rescue teams to better coordinate their efforts and identify areas that have already been searched.

\subsection{Disaster Management and Humanitarian Aid}
Some of the key ways that UAVs can be used in disaster management and humanitarian aid are as follows \cite{song2022toward, daud2022applications}:

\textbf{Damage Assessment:} UAVs can be used to quickly assess the extent of damage caused by disasters such as earthquakes, floods, and hurricanes. They can be equipped with high-resolution cameras and sensors to capture images and data of affected areas, which can be used to prioritize relief efforts.

\textbf{Monitoring and Surveillance:} UAVs can be used to monitor disaster-affected areas and detect any potential hazards or risks such as landslides, floods, or fires. They can also be used to track relief efforts and identify areas that still require assistance.

\section{Military and Defense applications}
Military and defense applications of UAVs include reconnaissance and surveillance, battle damage assessment, airborne communications and relays, combat and weaponized UAVs, and various other tasks where UAVs can provide improved situational awareness, reduced risk to human pilots, and enhanced capabilities in operations.

\subsection{Reconnaissance and Surveillance}
Some major reconnaissance and surveillance applications of UAVs are as follows \cite{ciolponea2022integration}:

\textbf{Intelligence, Surveillance, and Reconnaissance (ISR):} UAVs can be used for intelligence gathering, surveillance, and reconnaissance purposes. They can be equipped with high-resolution cameras, radar, and other sensors to capture images and data of enemy positions, movements, and activities. This information can be used to support military planning and decision-making.

\textbf{Border and Coastal Surveillance:} UAVs can be used for border and coastal surveillance to monitor illegal activities such as smuggling and trafficking. They can be equipped with thermal imaging sensors and other equipment to detect suspicious activities and identify potential threats.

\textbf{Battlefield Monitoring:} UAVs can be used for battlefield monitoring to track enemy movements and activities. They can be used to provide real-time information to military commanders, helping them to make informed decisions and adjust their tactics accordingly.

\textbf{Target Acquisition:} UAVs can be used for target acquisition to identify and locate enemy targets. They can be equipped with laser designators and other equipment to mark targets for attack by other military assets.

\textbf{Electronic Warfare:} UAVs can be used for electronic warfare purposes such as jamming enemy communications and disrupting their operations \cite{kratky2020electronic}. They can be equipped with electronic warfare equipment such as jammers and decoys to confuse and disrupt enemy systems.

\subsection{Battle Damage Assessment (BDA)}
AVs can be used in Battle Damage Assessment (BDA) applications to provide accurate and timely information on the effects of military operations \cite{cui2022application, kwak2022autonomous}.

\textbf{Post-Strike Damage Assessment:} UAVs can be used to assess the extent of damage caused by military strikes. They can be equipped with high-resolution cameras and sensors to capture images and data of the affected areas, which can be used to assess the effectiveness of the strikes and plan for follow-up operations.

\textbf{Target Verification:} UAVs can be used to verify that the intended targets have been hit during military operations. They can be equipped with high-resolution cameras and sensors to capture images and data of the targets before and after the strikes, which can be used to verify the effectiveness of the strikes.

\textbf{Damage Assessment Planning:} UAVs can be used to plan for damage assessment operations before military strikes. They can be used to map out the target area and identify potential risks and hazards, helping military commanders to plan for follow-up operations.

\subsection{Airborne Communications and Relays}
UAVs can be used in military applications as airborne communications and relays to provide secure and reliable communication links between ground forces and command centers.

\textbf{Signal Relay:} UAVs can be used to relay communication signals between ground forces and command centers. They can be equipped with communication equipment such as radios and satellite links to provide reliable and secure communication links over long distances.

\textbf{Network Extension:} UAVs can be used to extend the range of existing communication networks. They can be used to provide coverage in areas where communication infrastructure is limited or non-existent, such as remote areas or during natural disasters.

\textbf{Mobile Communication Centers:} UAVs can be used as mobile communication centers to provide on-demand communication support to ground forces. They can be deployed quickly to provide communication support in areas where infrastructure has been damaged or destroyed.

\textbf{Surveillance and Communications:} UAVs can be used to provide both surveillance and communication capabilities. They can be equipped with cameras and sensors to provide real-time information to military commanders, while also relaying communication signals between ground forces and command centers.

\textbf{Rapid Deployment:} UAVs can be rapidly deployed to provide communication support in emergency situations. They can be launched from remote locations and quickly reach the affected areas, providing critical communication links to ground forces.

\subsection{Combat and Weaponized UAVs}
As outlined below, there are some ways that UAVs can be used in combat and as weaponized platforms \cite{ren2022application, da2022criminal}.

\textbf{Offensive Operations:} UAVs can be used for offensive operations, including airstrikes and missile attacks. They can be equipped with weapons such as missiles, bombs, and guns to deliver precision strikes against enemy targets.

\textbf{Defensive Operations:} UAVs can be used for defensive operations, including air defense and counter-insurgency missions. They can be equipped with weapons such as anti-aircraft missiles and anti-tank missiles to protect military forces and assets.

\textbf{Swarm Tactics:} UAVs can be used in swarm tactics to overwhelm enemy defenses. They can be launched in large numbers to provide a coordinated attack against enemy forces.

Table I provides examples of common applications and benefits of UAVs in various industries and fields, including aerial photography, surveying and mapping, agriculture and forestry, infrastructure inspection and maintenance, search and rescue operations, military reconnaissance and surveillance, battle damage assessment, airborne communications and relays, and combat and weaponized UAVs.

\begin{table*}[hbt!!]
  \centering
  \caption{Civilian and Military Applications of UAVs: Examples of Applications and Benefits}
  \label{tab:mechanisms}
  \begin{tabular}{|p{40mm}|p{57mm}|p{57mm}|}
    
    \hline
    \textbf{Industry/Field} & \textbf{Application} & \textbf{Benefit}\\
    \hline

Aerial Photography/Videography	& Filming, photography, cinematography	& Access to difficult-to-reach locations, cost-effective, high-quality imaging \\
    \hline

Surveying and Mapping	& Topographic mapping, land surveying, 3D modeling	& Faster, safer, more accurate data collection, improved mapping capabilities \\
    \hline

Agriculture and Forestry	& Crop health monitoring, yield optimization, forest management	& Improved crop management, reduced costs, increased yields, better forest management \\
    \hline

Infrastructure Inspection and Maintenance	& Building and bridge inspections, pipeline monitoring, power line inspections	& Safer, faster, and more cost-effective inspections, improved maintenance \\
    \hline

Search and Rescue Operations		& Missing person search, disaster response, law enforcement support	& Improved situational awareness, faster response times, safer operations \\
    \hline

Military Reconnaissance and Surveillance	& Intelligence gathering, target tracking, border surveillance	& Improved situational awareness, reduced risk to human pilots \\
    \hline

Battle Damage Assessment	& Post-strike assessment, target identification	& Improved accuracy, reduced risk to human pilots \\
    \hline

Airborne Communications and Relays	& Communication support, data transmission	& Improved communication capabilities in remote areas, extended range \\
    \hline

Combat and Weaponized UAVs	& Combat operations, targeted strikes	& Reduced risk to human pilots, Improved accuracy and precision \\
    \hline

    \end{tabular}
\end{table*}

\section{Future of UAVs}
The future of UAVs is expected to see advancements in technology, new applications, and increased concerns about their impact on society.

\subsection{Advancements in UAV technology}
UAV technology has already come a long way in recent years, but there are several advancements that we can expect to see in the future \cite{ren2022application, bouguettaya2022deep, yu2022deep, gokbel2022improvement}.

\textbf{Improved Flight Performance:} One area of advancement is in flight performance. UAVs will likely become more agile and maneuverable, with longer flight times and greater range. Improvements in battery and propulsion technology may also lead to UAVs with higher speed and altitude capabilities.

\textbf{Artificial Intelligence (AI):} AI will play a significant role in future UAV technology. UAVs may become more autonomous, with the ability to make decisions on their own based on their environment and mission objectives. This will require advances in machine learning, computer vision, and sensor technology.

\textbf{Sensor Technology:} Sensor technology is another area of advancement in UAV technology. Future UAVs may be equipped with advanced sensors such as LiDAR, radar, and hyperspectral cameras, enabling them to perform a wider range of missions with greater accuracy.

\textbf{Communication Technology:} Communication technology will also play a critical role in future UAVs. UAVs may be equipped with advanced communication systems that enable them to communicate with each other, as well as with other aircraft and ground-based systems.

\textbf{Swarm Technology:} Swarm technology is another area of advancement in UAV technology \cite{kong2022research}. Future UAVs may be able to operate in coordinated groups, enabling them to perform complex missions and provide greater situational awareness to military commanders.

\textbf{Materials Science:} Advances in materials science will also play a role in future UAV technology. New materials such as lightweight composites and 3D printed components may enable the development of smaller and more efficient UAVs.

\subsection{Potential new applications of UAVs}
As UAV technology continues to advance, there are several potential new applications of UAVs that could emerge. 

\textbf{Urban Air Mobility:} One potential new application of UAVs is in the field of urban air mobility \cite{liberacki2023environmental, causa2023strategic}. UAVs may be used to transport people and goods within and between cities, helping to alleviate traffic congestion and reduce travel times.

\textbf{Delivery Services:} UAVs may also be used for delivery services, providing a faster and more efficient way to deliver packages and goods. This could have significant applications in fields such as e-commerce and emergency medical services.

\textbf{Space Exploration:} UAVs may also be used in space exploration missions, providing a low-cost and efficient way to explore other planets and celestial bodies \cite{akhil2022applications, sharma2022survey}. This could include the use of UAVs to collect samples or conduct reconnaissance missions.

\textbf{Environmental Monitoring:} UAVs may also be used for environmental monitoring purposes. UAVs could be equipped with sensors to measure things like air quality, temperature, and water quality, helping to provide real-time data for environmental monitoring and research.

\textbf{Wildlife Conservation:} UAVs may also be used in wildlife conservation efforts. UAVs could be used to monitor wildlife populations and track animal movements, helping to inform conservation efforts and protect endangered species.

\textbf{Construction and Maintenance:} UAVs may also be used for construction and maintenance purposes. UAVs could be used to perform inspections and surveys of buildings and infrastructure, as well as assist in construction and maintenance tasks.

\subsection{Concerns about UAVs} 
As UAV technology continues to advance and UAVs become more widely used, there are several concerns about their impact on society. 

\textbf{Privacy:} One concern is that UAVs could be used for surveillance purposes, potentially violating individuals' privacy rights \cite{lee2022safety}. This is particularly concerning in the context of law enforcement and government surveillance, where the use of UAVs could lead to overreach and abuse of power.

\textbf{Safety:} Another concern is the potential safety risks associated with UAVs. UAVs could collide with other aircraft, crash into buildings or other structures, or injure people on the ground if they malfunction or are operated improperly.

\textbf{Security:} There are also concerns about the security implications of UAVs \cite{pandey2022security}. UAVs could be used to carry out acts of terrorism, smuggle contraband into secure areas, or conduct corporate espionage.

\textbf{Noise Pollution:} UAVs can also produce a significant amount of noise pollution, particularly when operated in urban areas. This can be a nuisance to individuals and businesses in the area.

\textbf{Job Displacement:} The increasing use of UAVs could also lead to job displacement in certain industries, particularly in fields such as transportation and delivery \cite{marquez2022unemployment}.

\textbf{Environmental Impact:} UAVs could also have an environmental impact, particularly if they are powered by fossil fuels \cite{ccalicsir2023benchmarking}. The use of UAVs could contribute to air pollution and climate change.

\section{Concluding Remarks}
Rapid development of UAV technology has opened up a world of possibilities for a variety of industries and applications. The versatility, efficiency, and affordability of UAVs have made them an attractive option for a wide range of tasks, from aerial photography and surveying to search and rescue operations and military reconnaissance. The potential for future advancements in UAV technology is also promising, and could lead to even more applications and benefits. However, it is important to address concerns about the impact of UAVs on society, such as privacy, safety, and environmental impact. By taking a responsible and ethical approach to the use of UAVs, we can unlock their full potential and harness the benefits of this exciting technology for the betterment of society.

In the era of industry 4.0, digital technologies such as artificial intelligence, machine learning, and the Internet of Things are rapidly advancing and converging, creating new opportunities for UAV development \cite{hoque2021iotaas}. To progress in this environment, UAVs will need to become more autonomous, able to operate in complex and dynamic environments, and integrate with other digital systems seamlessly. This will require advancements in sensors, data processing capabilities, and communication networks. Additionally, the development of industry standards and regulations will be critical to ensure the safe and responsible use of UAVs in a connected world.

\bibliographystyle{IEEEtran}
\bibliography{reference}

\end{document}